\documentclass[runningheads]{llncs}
\usepackage{graphicx}
\usepackage{amsmath,amssymb} 
\usepackage{color}\usepackage[width=122mm,left=12mm,paperwidth=146mm,height=193mm,top=12mm,paperheight=217mm]{geometry}
\newcommand*\samethanks[1][\value{footnote}]{\footnotemark[#1]}
\usepackage{xcolor}
\usepackage{subfig}
\def\etal{et al. }

\begin{document}
\pagestyle{headings}
\mainmatter
\title{Online Detection of Action Start in Untrimmed, Streaming Videos} 

\titlerunning{title running}

\authorrunning{authors running}

\author{Zheng Shou\thanks{indicates equal contributions.}$^1$, Junting Pan\samethanks$^{1,2}$, Jonathan Chan$^1$, Kazuyuki Miyazawa$^3$, Hassan Mansour$^3$, Anthony Vetro$^3$, Xavi Giró-i-Nieto$^2$, Shih-Fu Chang$^1$}
\institute{$^1$Columbia University, $^2$Universitat Politecnica de Catalunya, $^3$Mitsubishi Electric}

\maketitle

\begin{abstract}
	We aim to tackle a novel task in action detection - Online Detection of Action Start (ODAS) in untrimmed, streaming videos. The goal of ODAS is to detect the start of an action instance, with high categorization accuracy and low detection latency. ODAS is important in many applications such as early alert generation to allow timely security or emergency response. We propose three novel methods to specifically address the challenges in training ODAS models: (1) hard negative samples generation based on Generative Adversarial Network (GAN) to distinguish ambiguous background, (2) explicitly modeling the temporal consistency between data around action start and data succeeding action start, and (3) adaptive sampling strategy to handle the scarcity of training data. We conduct extensive experiments using THUMOS'14 and ActivityNet. We show that our proposed methods lead to significant performance gains and improve the state-of-the-art methods. An ablation study confirms the effectiveness of each proposed method.
\end{abstract}


\section{Introduction}

In this paper, we investigate a novel task - \textbf{Online Detection of Action Start (ODAS)} in untrimmed, streaming videos:

\textbf{i}. \textbf{Online detection} requires continuously monitoring the live video stream in real time. When a new video frame arrives, online detection system processes it immediately, without any side information or access to the future frames.

\textbf{ii}. We refer the start/onset of an action instance as its \textbf{Action Start (AS)}, which is a time point and is associated with one action instance.

\textbf{iii}.
Following recent online detection works \cite{eccv16_oad,bmvc17_red}, we target \textbf{untrimmed, long, unconstrained} videos with large amounts of complex background streams.
ODAS differs from prior works on early event detection such as \cite{Hoai-DelaTorre-CVPR12,Hoai-DelaTorre-IJCV14}, which targeted relatively simple videos and assumed that each video contained only one action instance and which the action class is going to happen is known beforehand.
ODAS targets the more practical setting that each video can contain multiple action instances and the action class in the testing video is not known in advance.

\begin{figure}[t]
	\centering
	\includegraphics[width=.9\textwidth]{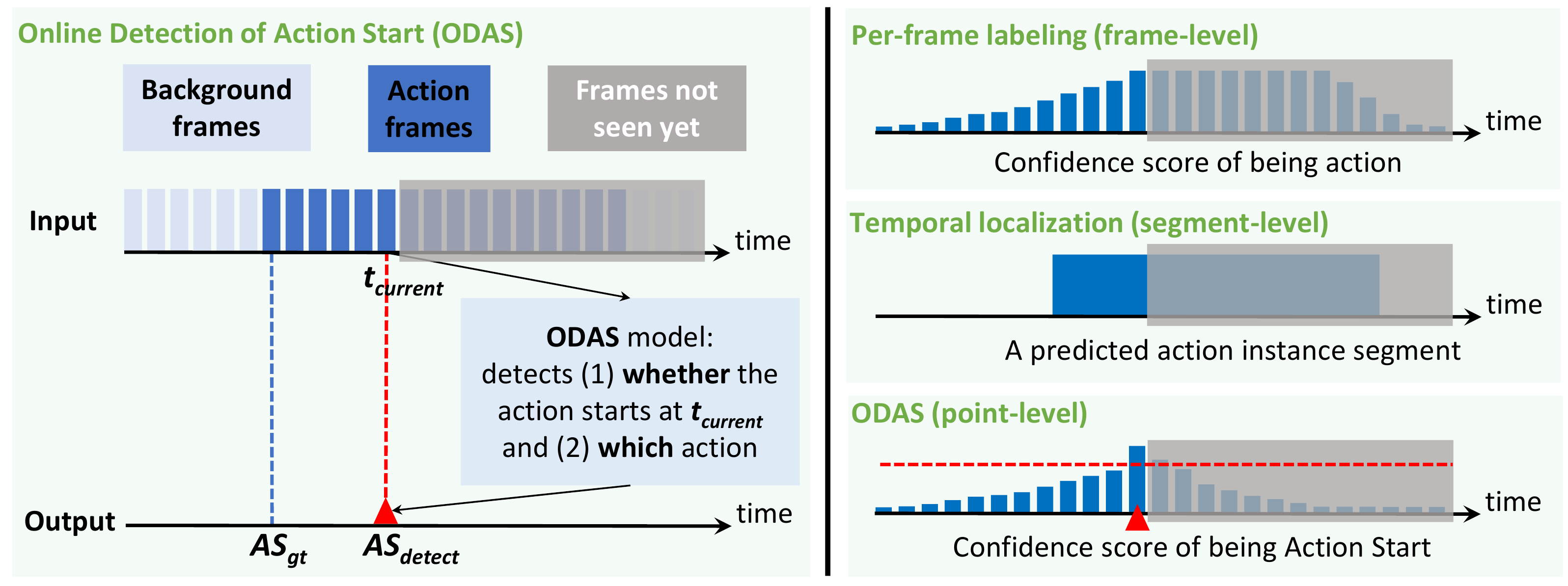}
	\caption{
		\textit{Left}: Illustration of the novel \textbf{Online Detection of Action Start (ODAS)} task; \textit{Right}: comparisons with the conventional action detection tasks
	}
	\label{intro_ODAS}
\end{figure}

As illustrated in Fig.~\ref{intro_ODAS} left, ODAS aims to detect the occurrence and class of AS as soon as the action happens.
ODAS is very important in many practical application scenarios, such as early alert generation.
For example,
the surveillance camera monitoring system needs to detect AS and then issue an alert as soon as possible to allow timely security response;
autonomous driving car needs to detect AS of accidents happening in front of it as soon as possible so that the car can slow down or change course timely to avoid collision;
robot looking after walking-impaired people shall detect AS of falling as soon as possible to provide assistance before the person has fallen down already.
Consequently,
in each of such scenarios, it is important to detect AS timely and accurately.

As illustrated in Fig.~\ref{intro_ODAS} right, ODAS differs from the conventional action detection tasks. 
Recent online detection works \cite{eccv16_oad,bmvc17_red} target the \textbf{per-frame labeling} task which aims to correctly classifying every frame into either the background class or certain action classes.
Recent offline detection works \cite{fast_temporal_activity_cvpr16,victor_eccv16,Richard_2016_CVPR,stanford_cvpr16,yuan_cvpr16,scnn_shou_wang_chang_cvpr16,cdc_zheng_cvpr17,lin2017single,heilbron2017scc,zhao2017temporal,iccv17_tap,xu2017r,dave2017predictive,dai2017temporal,bmvc17_tad,buch2017sst,sstad_buch_bmvc17,yuan2017temporal,sssn_mm15,sigurdsson2017asynchronous,bmvc17_st,iccv17_tall} target the \textbf{temporal localization} task which aims to predict a set of action segment instances in a long, untrimmed video.
Despite lacking the need to correctly classify every frame (as in per-frame labeling) or localize the complete action segment instance (as in temporal localization), ODAS explicitly focuses on detecting AS, which is quite challenging as discussed in the next paragraph. 
Traditional methods for per-frame labeling and temporal localization can indeed be adapted for ODAS.
But since they were originally designed to address different problems, the challenges in detecting AS have not been specifically considered and deeply investigated.
Therefore, methods excelling at per-frame labeling or temporal localization might not perform well in ODAS.

\begin{figure}[t]
	\centering
	\includegraphics[width=\textwidth]{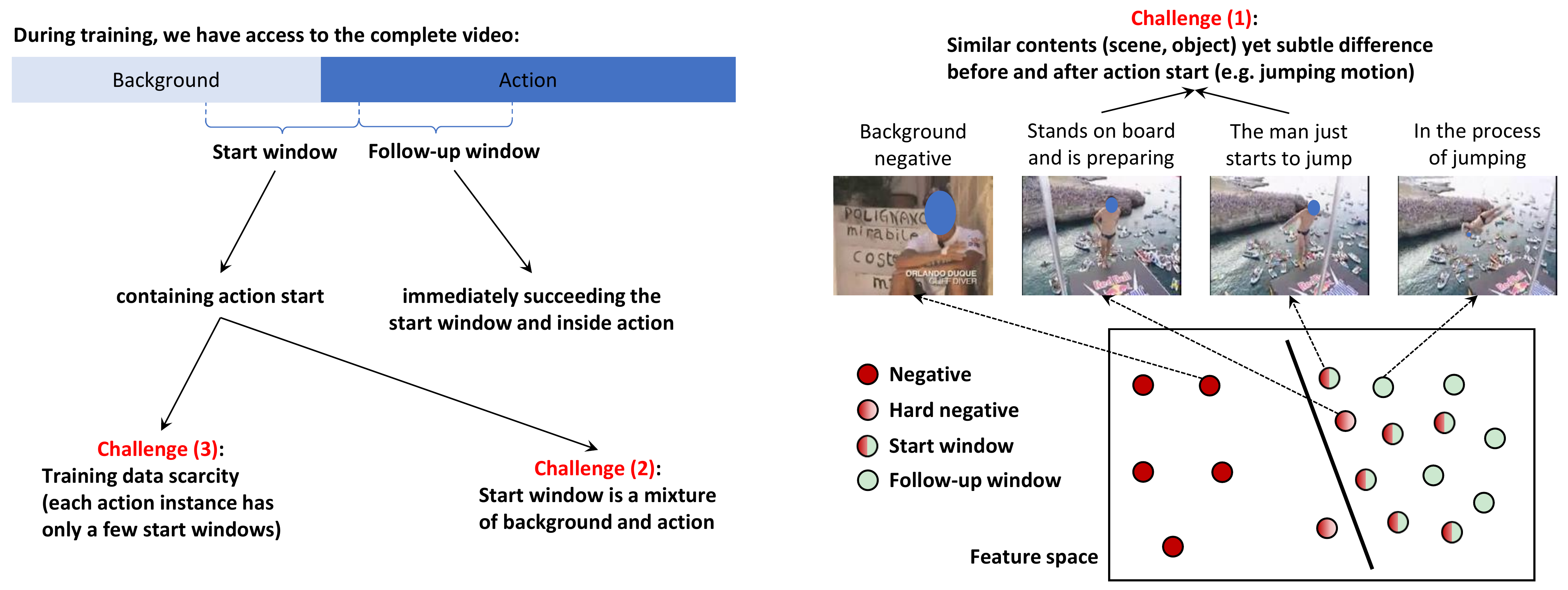}
	\caption{
		We identify three challenges in training a good ODAS model
	}
	\label{intro_ideas}
\end{figure}

In this paper, we identify three \textbf{challenges} in training a good ODAS model and accordingly propose three novel \textbf{solutions}.
\textcolor{red}{\textbf{(Challenge 1)}}
As the example shown in Fig.~\ref{intro_ideas}, it is important to learn and detect characteristics that can correctly distinguish the start window from the background, which precedes AS and may share very similar scenes but without the actual occurrence of actions.
Note that we follow the state-of-the-art video classification models such as C3D \cite{3dcnn,tran2017convnet}, TSN \cite{TSN}, I3D \cite{Kinetics} to accept short temporal sliding window as the network input.
To address this challenge, we introduce an auxiliary generative network trained in an adversarial process to automatically \textbf{generate hard negative samples} during training.
Although hard negative data may be rare in the training videos, our generator directly learns to model the distribution of hard negatives and thus can generate a much larger pool of hard negatives.
\textcolor{red}{\textbf{(Challenge 2)}}
We define the start window and its follow-up window in Fig.~\ref{intro_ideas}.
A start window contains both action frames and background frames.
Background preceding action can provide temporal contextual information but can also be confusing.
Due to the shared contents (background scene and object), the feature of the start window may be closer to the preceding background window than the actual action window after the start.
To remedy this issue, since the follow-up window is completely inside action, we propose to model the \textbf{temporal consistency} between the start window and its follow-up window during training.
\textcolor{red}{\textbf{(Challenge 3)}}
It is important to accurately classify start windows in ODAS.
But each action instance only has a few training samples of start windows, and thus the number of training samples for start windows is much more scarce than others such as background windows and windows fully inside action.
To address this issue, we design an \textbf{adaptive sampling} strategy to increase the percentage of start windows in each training batch.
Our experiments in Sec.~\ref{exp} will prove the effectiveness and necessity of each proposed method and putting three methods together results in significant performance gains.

In summary, we make three contributions in this paper:

(a) We propose a novel task ODAS in untrimmed, unconstrained, streaming videos to specifically focus on detecting AS timely and accurately.

(b) We design three novel methods for training effective ODAS models: (1) generating hard negative samples based on GAN to assist ODAS models in discriminating start windows from negatives, (2) modeling the temporal consistency between the start window and its follow-up window to encourage their feature similarity, and finally (3) adaptively sampling start windows more frequently to address the training sample unbalance issue.

(c) Extensive comparisons on THUMOS'14 and ActivityNet demonstrate the superiority of our approach over the conventional methods designed for online detection, per-frame labeling, temporal localization, and shot boundary detection in specifically solving the ODAS problem.

\section{Related Works} \label{relatedwork}

\subsubsection{Action Classification.}
Given a video clip, the goal of the classification task is to recognize the action categories contained in the whole video. Impressive progress \cite{TSN,Kinetics,dtf,idtf,FV1,VLAD1,3dcnn,Gan_2015_CVPR,Simonyan14b,Feichtenhofer_2016_CVPR,xu2015discriminative} has been made to address this problem.
Various network architectures have been proposed, such as 3D ConvNets, two-stream network \cite{Simonyan14b,TSN}, I3D \cite{Kinetics}, etc.
Detailed review can be found in surveys \cite{survey1,survey2,survey3,survey4,survey5,survey6}.

\subsubsection{Temporal Action Localization.}
Given a long, untrimmed video, temporal action localization needs to temporally localize each action instance: we not only predict its category but also detect when it starts and ends.
This task has raised a lot of interest in recent
\cite{fast_temporal_activity_cvpr16,victor_eccv16,Richard_2016_CVPR,stanford_cvpr16,yuan_cvpr16,scnn_shou_wang_chang_cvpr16,cdc_zheng_cvpr17,lin2017single,heilbron2017scc,zhao2017temporal,iccv17_tap,xu2017r,dave2017predictive,dai2017temporal,bmvc17_tad,buch2017sst,sstad_buch_bmvc17,yuan2017temporal,sssn_mm15,sigurdsson2017asynchronous,bmvc17_st,iccv17_tall}.
Shou et al. \cite{cdc_zheng_cvpr17} proposed a Convolutional-De-Convolutional (CDC) network to detect precise segment boundary;
Zhao et al. \cite{zhao2017temporal} presented the Structured Segment Network (SSN) to model the temporal structure of each action segment;
Buch et al. \cite{sstad_buch_bmvc17} designed an end-to-end system to stream the video and detect action segments ending at the current time.

Although these methods for temporal localization were originally designed for the offline setting, some of them can be adapted to conduct temporal localization in an online manner.
However, besides detecting AS, temporal localization requires detecting the Action End (AE) as well.
To this end, many methods have to wait for seeing AE in order to localize the action segment as a whole so that can determine AS, resulting in high latency.
Also, a good temporal localization method may excel at AE detection but perform poorly in AS detection (considering a detected segment that overlaps with the ground truth segment with IoU 0.7 and has the same AE as the ground truth).
ODAS focuses on AS specifically.

\subsubsection{Early Recognition and Detection.}
Similar to ODAS, early recognition and detection also aim to detect action as soon as it happens in streaming videos.
Early recognition was effectively formulated as partial action classification \cite{kong2014discriminative,yu2012predicting,cao2013recognize,ryoo2011human,kong2017deep,lan2014hierarchical,aliakbarian2017encouraging}:
the videos used in early recognition literatures are usually relatively short;
during testing, they cut each video to only keep its first certain portion of the whole video, and then classify the cut video into the pre-defined action classes.

ODAS is more related to early detection.
Hoai and De la Torre \cite{Hoai-DelaTorre-CVPR12,Hoai-DelaTorre-IJCV14} made attempts to detect actions in an online manner yet under a simplified setting (e.g., one action instance per video).
Huang et al. \cite{huang2014sequential} worked on a scenario that the background contents are simple (i.e. the person is standing and keeping still).
In this paper, like \cite{eccv16_oad,bmvc17_red}, we focus on the realistic videos that are unconstrained and contain complex backgrounds of large variety.
Ma et al. \cite{ma2016learning} approached the early detection task by cutting the first certain portion of the whole testing video and then conducting temporal localization on the cut video.
Hence, besides detecting AS, this work also focused on detecting whether the action ends or not.

\subsubsection{Online Action Detection.} 
Recent works on online action detection are very close to ODAS.
De Geest \etal \cite{eccv16_oad} first simulated the online action detection problem using untrimmed, realistic videos and benchmarked the existing models.
Gao \etal \cite{bmvc17_red} designed a training strategy to encourage a LSTM-based Reinforced Encoder-Decoder (RED) Network to make correct frame-level label predictions as early as possible.
But both of them formulated online action detection as online per-frame labeling task, which requires correctly classifying every frame rather than just detecting AS.
A good per-frame labeling method might not be necessarily good at detecting AS (considering a per-frame labeling method correctly classifying frames in the 30\%-100\% portion of each action instance but mis-classifying frames in its 0\%-30\% portion).
Consequently, as for the applications that detecting AS is the most important task, ODAS is the best fit.

In addition, there are also works on spatio-temporally localizing actions in an online manner but also limited to short videos \cite{singh2016online,soomro2016predicting}.
Li \etal \cite{li2016online} and Liu \etal \cite{liu2017online} leveraged Kinect sensors and performed detection based on the tracked skeleton information.
Vondrick \etal \cite{vondrick2015anticipating} targeted future prediction, which is a more ambitious online detection task.

\hfill\break\noindent\textbf{Adversarial Learning.}
The idea of training in an adversarial process was first proposed in \cite{goodfellow2014generative} and has been adopted in many applications
\cite{odena2016conditional,vondrick2017generating,isola2016image,zhu2017unpaired,tzeng2017adversarial}.
Generative Adversarial Network (GAN) \cite{goodfellow2014generative,radford2015unsupervised} consists of two networks trained simultaneously to compete with each other:
a generator network G that learns to generate fake samples indistinguishable from real data
and a discriminator network D which is optimized to recognize whether input data samples are real or fake.
To the best of our knowledge, we are the first to explore GAN for action detection.


\section{Framework} \label{framework}

In this Section, we introduce our ODAS framework as shown in Fig.~\ref{intro_ODAS}.
We follow the state-of-the-art video classification networks like C3D \cite{3dcnn,tran2017convnet}, TSN \cite{TSN}, I3D \cite{Kinetics} to accept temporal sliding windows as input.
In particular, we set the window length to 16 frames and use C3D as our backbone network in Sec.~\ref{framework} and Sec.~\ref{model} to help illustrate technical ideas.

We outline our ODAS framework by walking through the testing pipeline.
During testing, when a new frame arrives at the current time $t$, we immediately feed the streaming window ending at $t$ into our network.
The network output at $t$ consists of the semantic class $c_t$ which could be either background or action $1, \ldots ,K$ ($K$ is the total number of action classes) and the confidence score $s_t$.
In order to detect AS, we compare the network outputs at $t-1$ and $t$.
We generate an AS point prediction whenever the following conditions are all satisfied: (1) $c_t$ is action; (2) $c_t \neq c_{t - 1}$; (3) $s_t$ exceeds the threshold obtained by grid search on the training set.
Such an AS point prediction is associated with the time point $t$, the predicted class (set to $c_t$) and the confidence score (set to $s_t$).

As alternatives, we have also studied the approach of adding a proposal stage specifically for detecting action start and then classifying the action class.
We found such an alternative approach is not as effective as the one outlined above.
Quantitative comparisons can be found in the supplementary material.

\section{Our Methods} \label{model}

As for training our ODAS model, the complete videos are available during training. We slide windows over time with a stride of 1 frame to first construct a set of training windows to be fed into the network.
For each window, we assign its label as the action class of the last frame of the window.
In this section, we propose three novel methods to improve the capability of the backbone networks at detecting action in a timely manner.
We first illustrate our intuition of designing these methods and then present their formulations.
We close this section by summarizing the full objective used for training.
Implementation details can be found in the supplementary material.

\begin{figure}[t]
	\centering
	\includegraphics[width=.85\textwidth]{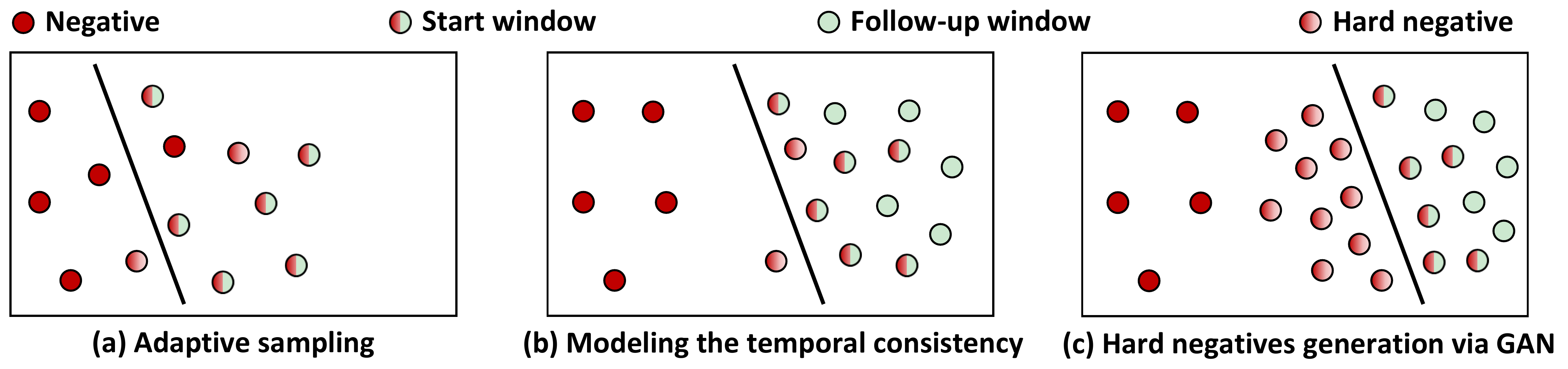}
	\caption{
		The effects of our proposed three training methods for ODAS. Each rectangle represents data distribution in the high-level feature space after training
	}
	\label{featspace}
\end{figure}

\begin{figure}[t]
	\centering
	\includegraphics[width=\textwidth]{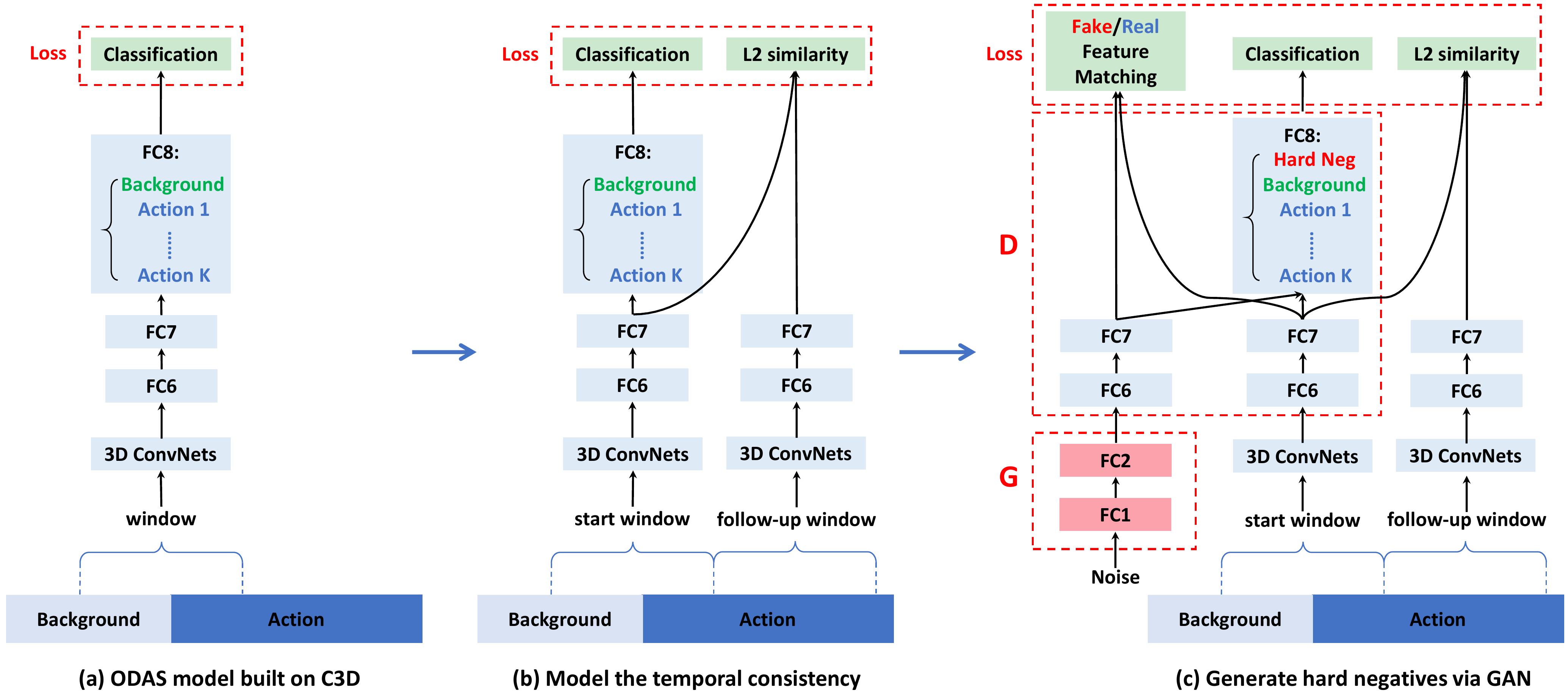}
	\caption{The network architectures of our ODAS models built on C3D \cite{3dcnn} and the proposed training objectives.
		(a) Our basic ODAS model consists of 3D ConvNets from $ \tt Conv1$ to $\tt Pool5$ and 3 fully connected layers ($ \tt FC6$, $ \tt FC7$, $ \tt FC8$).
		We keep the same backbone network architecture as C3D while setting the number of nodes in $ \tt FC8$ to $K+1$, standing for $K$ actions and background.
		The output of $ \tt FC8$ is used for calculating multi-class classification softmax loss.
		(b) We model the temporal consistency between the start window and its paired follow-up window by adding a temporal consistency loss term to minimize the L2 similarity computed using their $ \tt FC7$ activations.
		Two streams in this siamese network share the same parameters.
		(c) Further, we design a GAN-based framework to automatically generate hard negative samples to help our model more accurately distinguish actions against negatives. G is generator and D is discriminator. G accepts random noise as input and output fake $\tt Pool5$ features. We add an additional class in $ \tt FC8$ for fake samples. All blue blocks of the same name are the same layer and share weights. More details of GAN can be found in Section \ref{ganmodel}
	}
	\label{network}
\end{figure}

\subsection{Intuition}

\subsubsection{Adaptively sample the training data.}
Since we want to detect actions as soon as possible, it is important for ODAS to accurately classify start windows.
This is a challenging task because the start window contains various background contents, and the number of start windows is quite scarce.
If we construct each training batch via randomly sampling out of all windows, the model might not see enough start windows during training in order to generalize to the testing data well.
To solve this issue, we guide the ODAS model to pay more attention to start windows via adaptively sampling more start windows during each training batch.
Using this strategy, the ODAS model can learn a better classification boundary to distinguish start windows against negatives more accurately, as illustrated in Fig.~\ref{featspace} (a).

\subsubsection{Model the temporal consistency.}
As illustrated in Fig.~\ref{featspace} (a), the start window is a mixture of action frames and background frames.
Therefore, in the feature space, start windows could be close to or even mixed with negatives.
It is important to accurately distinguish start windows and negatives in ODAS so that the model can more timely detect action start when the video stream switches from negative to action.
As illustrated in Fig.~\ref{featspace} (b), since the follow-up windows are completely inside action, they are far away from negative data in the feature space.
Thus we explicitly model the \textbf{Temporal Consistency (TC)} between each start window and its follow-up window to encourage their feature similarity.
Using this training method, start windows move closer to follow-up windows and thus become more separable from negatives.

\subsubsection{Generate hard negative samples via GAN.}
As exampled in Fig.~\ref{intro_ideas}, it is important to train the ODAS model to capture the subtle differences that can serve as evidences for discriminating start windows from negatives preceding AS.
As illustrated in Fig.~\ref{featspace} (b), hard negatives that have subtle differences with start windows might be close to start windows in the feature space.
In order to learn a better decision boundary, we aim to identify such hard negative samples in the training set during training.
However, exhaustively finding such samples is time-consuming because such hard negatives are rare and may even not exist in the training data.
Therefore, we propose to train a model to automatically synthesize samples that are hard to distinguish from true start windows.
As illustrated in Fig.~\ref{featspace} (c), equipped by these synthesized hard negatives, we can learn an even better ODAS model to discriminate start windows from negatives.

\subsection{Adaptively Sample the Training Data}

Concretely, we randomly sample half of the training batch from start windows and randomly sample the other half batch from the remaining windows, which can be backgrounds or windows completely inside actions.
After each training batch is constructed, we can feed them into our network as shown in Fig.~\ref{network} (a) and train the network via minimizing the multi-class classification softmax loss ${{\cal L}_{{\rm{classification}}}}$.
We denote the set of start windows as ${p_{{\rm{start}}}} = \left\{ {\left( {x_{s},y} \right)} \right\}$ where $x_{s}$ is the start window to be fed into our model and $y$ is its corresponding ground truth label.
Similarly, we express the set of remaining windows as ${p_{{\rm{notstart}}}} = \left\{ {\left( { {x_{ns}},y} \right)} \right\}$.
The label space of $y$ is $1, \ldots ,K+1$ where the first $K$ classes are actions and the $K+1$-th class stands for background.
Our network takes $x$ as input and predicts a vector $\left\{ {{o_1}, \ldots ,{o_{K + 1}}} \right\}$ of $K+1$ dimension.
Finally we apply the softmax function to obtain the normalized probability of being class $i$:
${P_{{\rm{model}}}}\left( {i|x} \right) = \frac{{{e^{{o_i}}}}}{{\sum\nolimits_{k = 1}^{K + 1} {{e^{{o_k}}}} }}$.
We use $\mathbb{E} \left[ \cdot \right]$ to represent expectation.
The classification loss is defined as:
\begin{equation}
\label{eq-adaptive}
\begin{array}{*{20}{c}}
{{{\cal L}_{{\rm{classification}}}} = }&{{\mathbb{E}_{\left( {x_{s},y} \right) \sim {p_{{\rm{start}}}}}}\left[ { - \log \left( {{P_{{\rm{model}}}}\left( {y|x_{s}} \right)} \right)} \right]}\\
{}&{ + {\mathbb{E}_{\left( { {x_{ns}} ,y} \right) \sim {p_{{\rm{notstart}}}}}}\left[ { - \log \left( {{P_{{\rm{model}}}}\left( {y| {x_{ns}} } \right)} \right)} \right]}.
\end{array}
\end{equation}

\subsection{Model the Temporal Consistency}

Formally, we denote the training set of the paired start window and follow-up window as ${p_{{\rm{startfollowup}}}} = \left\{ {\left( {x_{s},x_{f},y} \right)} \right\}$,
where $x_{s}$ represents the start window and $x_{f}$ is its associated follow-up window and $y$ is still the ground truth label.
We model the temporal consistency via minimizing the similarity ${\cal L}_{{\rm{similarity}}}$ measured by L2 distance of the feature representation between $x_{s}$ and $x_{f}$:
\begin{equation}
\label{eq-tc}
{{\cal L}_{{\rm{similarity}}}} = {\mathbb{E}_{\left( {x_{s},x_{f},y} \right) \sim {p_{{\rm{startfollowup}}}}}}\left\| {F\left( {x_{s}} \right) - F\left( {x_{f}} \right)} \right\|_2^2, 
\end{equation}
where the function $F\left(  \cdot  \right)$ indicates extracting feature representation.
As shown in Fig.~\ref{network} (b), we set $F\left(  \cdot  \right)$ to be the output of $\tt FC7$ because it is also the input to the final layer $\tt FC8$ for classification.
Trained with this temporal consistency loss, the features of start windows become more distinguishable from negatives, which leads to the ODAS performance improvements as shown in Sec.~\ref{ablation}.


\subsection{Generate Hard Negative Samples via GAN} \label{ganmodel}

In order to separate start windows and hard negatives which have subtle differences from start windows, we design a GAN-based framework which synthesizes hard negative samples to assist ODAS model training.

\subsubsection{Generator (G).} Since directly generating video is very challenging, we use GAN to generate features rather than raw videos.
As shown in Fig.~\ref{network} (c), our GAN model has a fixed 3D ConvNets (from $\tt Conv1$ to $\tt Pool5$) to extract real $\tt Pool5$ features from raw videos and also has a G to generate fake $\tt Pool5$ features. Upper layers serve as \textbf{Discriminator (D)}, which will be explained later.

G accepts a random noise $z$ as input and learns to capture the true distribution of real start windows.
Consequently, G has the potential to generate various fake $\tt Pool5$ samples which may not exist in the real training set but may appear during testing.
This enables our model to continuously explore more discriminative classification boundary in the high-level feature space.
Following \cite{radford2015unsupervised}, $z$ is a 100-dimensional vector randomly drawn from the standard normal distribution.
In practice, we find that a simple G consisting of two fully connected layers $\tt FC1$ and $\tt FC2$ works well. Each fully connected layer is followed by a BatchNorm layer and a ReLU layer.

When training G, the principle is to generate \textbf{hard} negative samples that are similar to real start windows.
Conventional GANs utilize a binary real/fake classifier to provide supervision signal for training G.
However, this method usually encounters an instability issue.
Following \cite{salimans2016improved}, instead of adding a binary classifier, we require G to generate fake data matching the statistics of the real data.
Specifically, the feature matching objective is forcing G to match the expectation of the real features on an intermediate layer of D (we use $\tt FC7$ layer as indicated in Fig.~\ref{network} (c)).
Formally, we denote the feature extraction part of using the fixed 3D ConvNets as $\phi \left( \cdot  \right)$ and the process from $\tt Pool5$ to $\tt FC7$ as $\psi \left(  \cdot  \right)$. The feature matching loss is defined as follows:
\begin{equation}
\label{eq-matching}
{{\cal L}_{\rm{matching}}} = \left\| {{\mathbb{E}_{\left( {x_{s},y} \right) \sim {p_{{\rm{start}}}}}}\left[ {\psi \left( {\phi \left( {{x_{s}}} \right)} \right)} \right] - {\mathbb{E}_{z \sim noise}}\left[ {\psi \left( {G\left( z \right)} \right)} \right]} \right\|_2^2,
\end{equation}
where $G\left(  \cdot  \right)$ denotes the generator, ${p_{{\rm{start}}}} = \left\{ {\left( {x_{s},y} \right)} \right\}$ is the training set of start windows, $x_{s}$ represents the start window, and $y$ is the ground truth label.

\subsubsection{Discriminator (D).} 
The principle for designing D is that the generated samples should be still separable from real start windows despite their similarity, so that the generated samples can be regarded as hard \textbf{negatives}.
As shown in Fig.~\ref{network} (c), D consists of $\tt FC6$, $\tt FC7$, and $\tt FC8$. Instead of adding a binary real/fake classifier, we add an additional node in $\tt FC8$ layer to represent the hard negative class, which is the ground truth label for the generated samples.
Note that this additional class is used during training only and is removed during testing.

Similarly, some previous works also replaced the binary discriminator with a multi-class classifier that has an additional class for the fake samples \cite{salimans2016improved,springenberg2015unsupervised,nips17_badgan_zhilin}. However, their motivation is mainly extending GAN to the semi-supervised setting: the unlabeled real samples could belong to any class except fake.
But in this paper, we focus on generating hard negatives which should be similar to actions but dissimilar to backgrounds; meanwhile our D needs to distinguish hard negatives from not only actions but also from backgrounds.

Given a $\tt Pool5$ feature ${\phi \left( {x} \right)}$ either extracted from real data or generated by G, D accepts ${\phi \left( {x} \right)}$ as input and predicts a vector $\left\{ {{o_1}, \ldots ,{o_{K + 2}}} \right\}$ which goes through a softmax function to get class probabilities:
${P_{{\rm{D}}}}\left( {i|{\phi \left( {x} \right)}} \right) = \frac{{{e^{{o_i}}}}}{{\sum\nolimits_{k = 1}^{K + 2} {{e^{{o_k}}}} }}$, where $i \in \left\{ {1, \ldots ,K + 2} \right\}$.
Regarding the real samples, we can calculate their corresponding classification loss ${{\cal L}_{{\rm{real}}}}$ via extending ${{\cal L}_{{\rm{classification}}}}$ defined in Eq.~\ref{eq-adaptive}:
\begin{equation}
\label{eq-real}
\begin{array}{*{20}{c}}
{{{\cal L}_{\rm{real}}} = }&{{\mathbb{E}_{\left( {x_{s},y} \right) \sim {P_{{\rm{start}}}}}}\left[ -  {\log {P_D}\left( {y|\phi \left( {x_{s}} \right)} \right)} \right]}\\
{}&{+ {\mathbb{E}_{\left( { {x_{ns}},y} \right) \sim {P_{{\rm{notstart}}}}}}\left[ -  {\log {P_D}\left( {y|\phi \left( { {x_{ns}}} \right)} \right)} \right]}.
\end{array}
\end{equation}
As for the generated fake samples, the loss is:
\begin{equation}
\label{eq-fake}
{{\cal L}_{\rm{fake}}} = {\mathbb{E}_{z \sim noise}}\left[-  {\log {P_D}\left( {K + 2|G\left( z \right)} \right)} \right],
\end{equation}
where $K+2$ represents the hard negative class.

\subsection{The Full Objective}

We first pre-train the whole network via minimizing ${{\cal L}_{{\rm{classification}}}} + \lambda  \cdot {{\cal L}_{{\rm{similarity}}}},$
which combines the classification loss (Eq.~\ref{eq-adaptive}) and the temporal consistency loss (Eq.~\ref{eq-tc}) together with the weighting parameter $\lambda$.
Based on such initialization, we train G and D in an alternating manner during each iteration:
\textbf{When training G}, we fix D and train G so that the full objective is $\mathop {\min }\limits_G {{\cal L}_{\rm{G}}}$.
${{\cal L}_{\rm{G}}}$ contains only the feature matching loss (Eq.~\ref{eq-matching}): ${{\cal L}_{\rm{G}}} = {{\cal L}_{\rm{matching}}}$.
\textbf{When training D}, we fix G and train D so that the full objective is $\mathop {\min }\limits_D {{\cal L}_{\rm{D}}}$.
${{\cal L}_{\rm{D}}}$ contains the classification loss for both the real and fake samples (Eq.~\ref{eq-real} and Eq.~\ref{eq-fake}) and also the temporal consistency loss (Eq.~\ref{eq-tc}):
${{\cal L}_{\rm{D}}} = {{\cal L}_{\rm{real}}} + {{\cal L}_{\rm{fake}}} + \lambda  \cdot {{\cal L}_{{\rm{similarity}}}}$,
where $\lambda$ is the weight.

\section{Evaluation} \label{eval}

\subsection{Conventional Protocols and Metrics}

Hoai and De la Torre \cite{Hoai-DelaTorre-CVPR12,Hoai-DelaTorre-IJCV14} first worked on early detection and proposed three evaluation protocols to respectively evaluate classification accuracy, detection timeliness, and localization precision.
As comprehensively discussed in \cite{eccv16_oad}, their protocols do not suit online detection in realistic, unconstrained videos, because their protocols were designed for a simplified setting: each video contains only one action instance of interest.

Therefore, as mentioned in Sec.~\ref{relatedwork}, recent online detection works \cite{eccv16_oad,bmvc17_red} effectively worked on the per-frame labeling task and evaluated the frame-level classification mean Average Precision (mAP) or its calibrated version.
In addition, temporal localization methods detect both the start and end times of each action instance and evaluate the segment-level detection mAP.
As for ODAS, the performance of detecting AS can indeed affect both the frame-level mAP and the segment-level mAP.
However, since the frame-level mAP is mainly used to evaluate the accuracy of classifying every frame and the segment-level mAP involves evaluating the correctness of detecting the end, both metrics are not exactly evaluating the performance in detecting action starts. 
Detailed discussions and examples can be found in the supplementary material.

\subsection{Proposed New Protocol and Metrics}



In order to specifically evaluate ODAS performance, we propose a new evaluation protocol.
We evaluate ODAS at the point level for each AS instance.
As mentioned in Sec.~\ref{framework}, ODAS system outputs a rank list of detected AS points. 
Each AS point prediction is associated with the time point $t$, the predicted action class and the confidence score.

We aim to evaluate the detection accuracy and timeliness of ODAS system compared to the human annotated ground truths \protect\footnotemark \footnotetext{According to the human annotations on THUMOS'14 test set \cite{THUMOS14}, people usually can form agreement on when the action starts: out of 3,358 action instances, 3,259 instances (97\%) have their AS time annotations agreed by multiple human annotators. The instances with ambiguous start times are excluded during evaluation.}.
As for the timeliness, inspired by the segment-level mAP which measures the temporal overlap between the ground truth segment and the predicted segment, we measures the temporal offset (absolute distance) between the ground truth AS point and the predicted AS point.
We discuss the temporal offset in detail in the supplementary material.

We propose the \textbf{point-level AS detection mAP} to evaluate ODAS results. Each AS point prediction is counted as correct only when its action class is correct and its offset is smaller than the evaluation threshold.
We evaluate the point-level AP for each action class and average over all action classes to get the point-level mAP.
We do not allow duplicate detections for the same ground truth AS point.

\section{Experiments} \label{exp}

In order to simulate the ODAS setting, we employ standard benchmarks consisting of untrimmed videos to simulate the sequential arrival of video frames.

\subsection{Results on THUMOS'14}


\subsubsection{Dataset.}
THUMOS'14 \cite{THUMOS14} involves 20 actions and videos over 20 hours:
200 validation videos (3,007 action instances) and 213 test videos (3,358 action instances).
These videos are untrimmed and contain at least one action instance.
Each video has 16.8 action instances in average.
We use the validation videos for training and use the test videos to simulate streaming videos for testing ODAS.

\subsubsection{Metrics.}
We evaluate the point-level AS detection mAP at different temporal offset thresholds to compare ODAS systems under the different user tolerances or application requirements.
Note that \textit{AP depth at recall X\%} means averaging the precisions at points on the P-R curve with the recall ranging from 0\% to X\%.
The aforementioned default detection mAP is evaluated at AP depth at recall 100\%.
In order to look at the precision of top ranked predictions, we can evaluate detection mAP at different AP depth.
At each AP depth, we average the detection mAP under different offset thresholds to obtain the average mAP.

\subsubsection{Comparisons.} As for \textcolor{cyan}{\textbf{our approach}},
our network architecture can be found in Fig.~\ref{network}.
Since we build our model upon C3D \cite{3dcnn} which has been pre-trained on Sports-1M \cite{sports1m}, we use it to initialize models shown in Fig.~\ref{network}.
Since $\tt Pool5$ output has 8,192 dimensions, we use 4,096 nodes for both $\tt FC1$ and $\tt FC2$ in G.
We compare with the following baselines.
(1) \textbf{Random guess}: we replace the network output scores for $K$ actions and background mentioned in Sec.~\ref{framework} via randomly splitting score 1 into $K+1$ numbers all within $\left[ {0,1} \right]$.
(2) \textcolor{brown}{\textbf{C3D w/o ours}}: we use the C3D model which has exactly the same network architecture as our model used during testing but is trained without our proposed methods.
(3) \textcolor{red}{\textbf{RED}}: Gao \etal \cite{bmvc17_red} achieved the state-of-the-art performances on THUMOS'14 in online action detection task by encouraging the LSTM network to make correct frame-level predictions at the early part of a sequence.
We obtained the results from the authors and evaluated based on our proposed protocol.
(4) Per-frame labeling method - \textbf{CDC}: Shou \etal \cite{cdc_zheng_cvpr17} designed a Convolutional-De-Convolutional Network to operate on the testing video in an online manner to output the per-frame classification scores, which can be used to determine AS points following the same pipeline as proposed in Sec.~\ref{framework}.
(5) Temporal localization method - \textbf{TAG in SSN}: Zhao \etal \cite{zhao2017temporal} proposed an effective segment proposal method called temporal actionness grouping (TAG).
Based on the actionness score sequence, TAG can be operated in an online manner to detect the start of a segment and then also detect its end.
Thus TAG can be used for ODAS to generate class-agnostic AS point proposals.
For fair comparisons, we determine the action score of each proposal by applying the AS classifier obtained by our best model (trained with all three methods).
(6) Shot boundary detection methods \cite{boreczky1996comparison,warhade2011video,smeaton2010video} detect the change boundaries in the video which can be considered as AS proposals. Then we utilize our best classifier to classify each AS proposal.
We employ two popular open-source shot detection methods \textbf{ShotDetect}\protect\footnotemark \footnotetext{https://github.com/johmathe/Shotdetect} and \textbf{SceneDetect}\protect\footnotemark \footnotetext{https://github.com/Breakthrough/PySceneDetect} respectively for comparisons.
(7) Offline detection method: \textbf{S-CNN} \cite{scnn_shou_wang_chang_cvpr16} uses the same C3D network architecture as our model but performs testing in offline.

\begin{figure}[t]
	\centering
	\subfloat[Comparisons with SoA]{{\includegraphics[height=0.19\textheight]{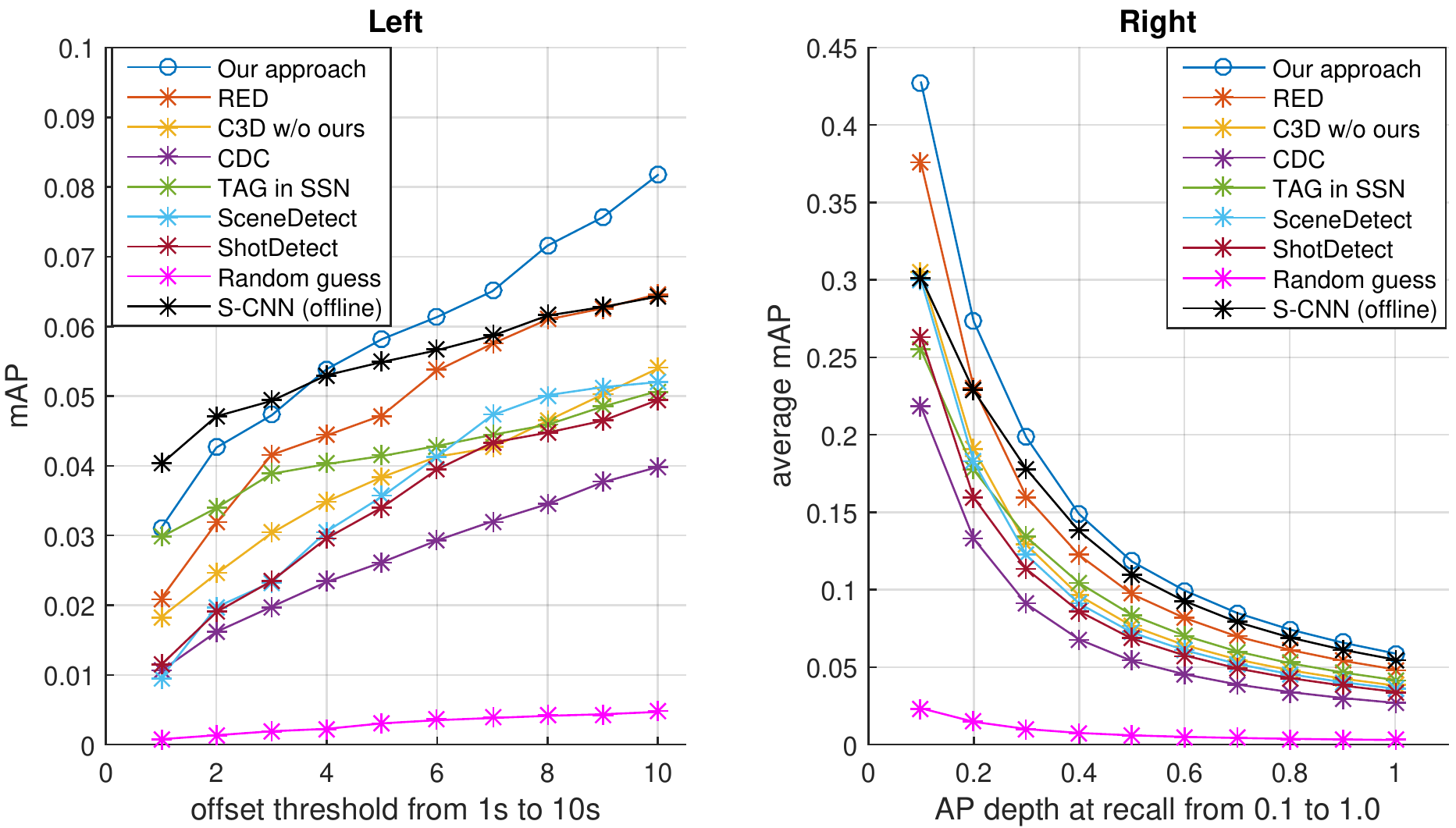} }}%
	\subfloat[Ablation study]{{\includegraphics[height=0.19\textheight]{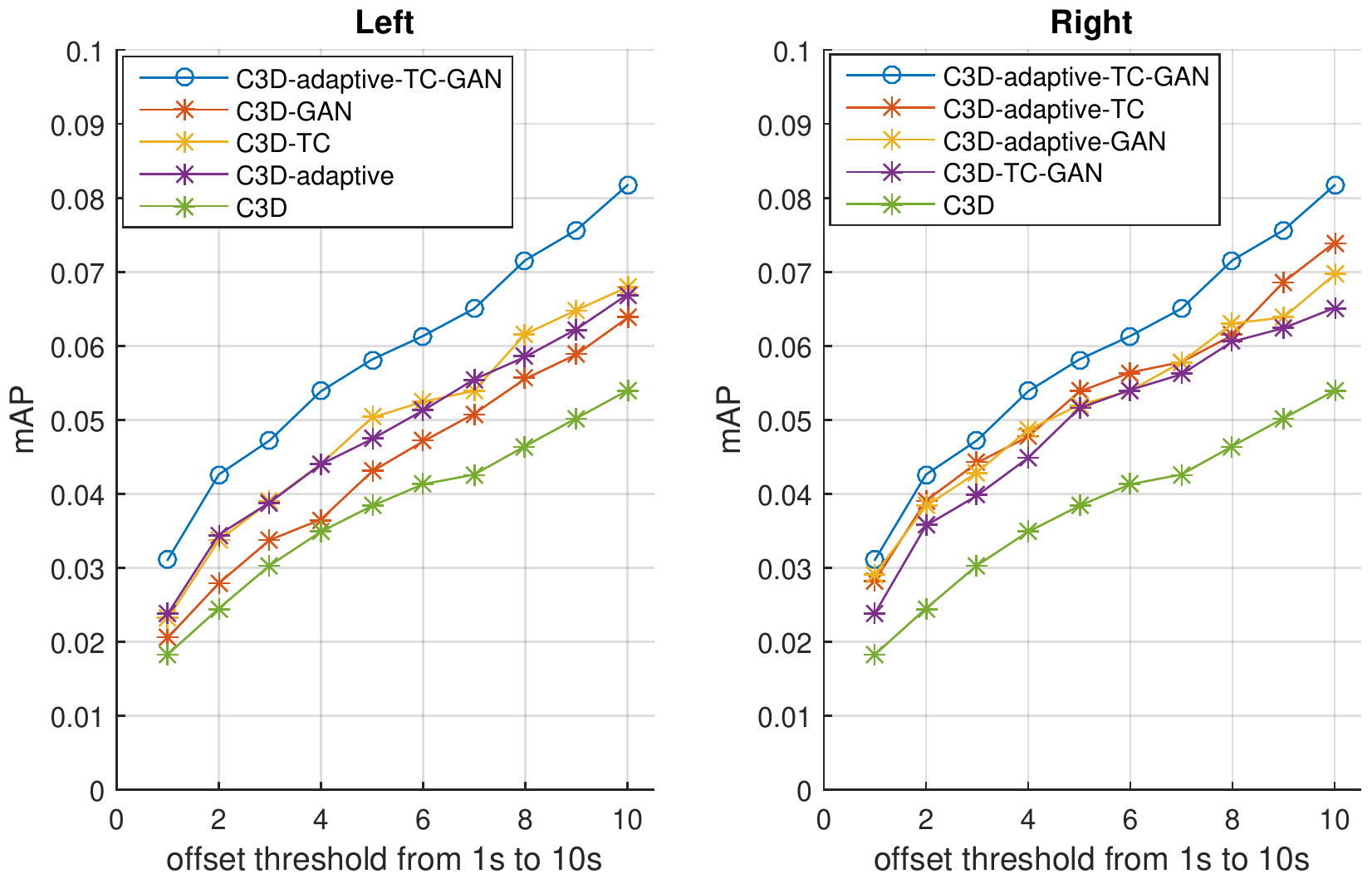} }}%
	\caption{Experimental results on THUMOS'14.
		(a) Comparisons with state-of-the-art methods. \textit{\textbf{Left}}: y-axis is the point-level AS detection mAP and x-axis is varying the offset threshold. \textit{\textbf{Right}}: y-axis is the average mAP averaged over offsets from 1s to 10s at AP depth at X\% recall and x-axis is varying X\% from 0.1 to 1.
		(b) Ablation study of our methods. Point-level AS detection mAP vs. offset threshold. \textit{\textbf{Left}}: Using one of proposed training methods only. \textit{\textbf{Right}}: removing one method out of the whole three combination
	}
	\label{res_th}
\end{figure}

Since the duration of action instance varies from $<$1s to $>$20s and thus during evaluation we vary the offset threshold from 1s to 10s. As shown in Fig.~\ref{res_th}, 
when using the proposed training strategies specifically designed to tackle ODAS, \textcolor{cyan}{\textbf{our approach}} improves the \textcolor{brown}{\textbf{C3D w/o ours}} baseline by a large margin.
Qualitative comparisons can be found in the supplementary material.
Notably, \textcolor{cyan}{\textbf{our approach}} is far better than \textbf{random guess} and also outperforms \textcolor{red}{\textbf{RED}}, which is a state-of-the-art method developed very recently to detect action as soon as possible.
Also, our method is better than other per-frame labeling method (i.e. \textbf{CDC}) and some class-agnostic start point detection methods (i.e. \textbf{TAG in SSN}, \textbf{ShotDetect}, \textbf{SceneDetect}).
Furthermore, \textcolor{cyan}{\textbf{our approach}} can even achieve better results under high offset thresholds than the offline \textbf{S-CNN} method.

\subsubsection{Ablation study of individual proposed method.} \label{ablation}
We conduct in-depth study on THUMOS'14 to analyze the performance gain contributed by each proposed training method.
In Fig.~\ref{res_th} (b) left, we report the results on THUMOS'14 when training with only one of our proposed methods. All approaches in the following have the same network architecture during testing:
\textbf{C3D} is trained without any proposed methods;
\textbf{C3D-adaptive} is trained with the adaptive sampling strategy;
\textbf{C3D-TC} is trained with modeling the temporal consistency;
\textbf{C3D-GAN} is trained within our proposed GAN-based framework;
\textbf{C3D-adaptive-TC-GAN} combines all three proposed methods together during training and achieves the best performance.
These results indicate that all three proposed methods are effective in improving the ODAS model.

In Fig.~\ref{res_th} (b) right, we report the results on THUMOS'14 when training without one of our proposed methods. We add additional approaches of the same network architecture during testing:
\textbf{C3D-TC-GAN} is trained without the adaptive sampling strategy;
\textbf{C3D-adaptive-GAN} is trained without modeling the temporal consistency;
\textbf{C3D-adaptive-TC} is trained without our proposed GAN-based framework.
These results indicate that all three proposed methods are necessary for training a good ODAS model.

\subsection{Results on ActivityNet}

\subsubsection{Dataset.} ActivityNet \cite{caba2015activitynet,activitynet} v1.3 involves 200 actions and untrimmed videos over 800 hours: around 10K training videos (15K instances) and 5K validation videos (7.6K instances).
Each video has 1.7 action instances in average.
We train on the training videos and evaluate ODAS using the validation videos.

\subsubsection{Comparisons.} As for our approach,
given the superior performances of TSN on ActivityNet video classification task \cite{TSN}, following \cite{bmvc17_tad}, for each window of 16 frames, we use TSN to extract a feature vector of 3,072 dimensions to serve as input to our network.
Our backbone network for ActivityNet consists of three fully connected layers (i.e. $\tt FC6$, $\tt FC7$, $\tt FC8$) that are the same as these in C3D, but we train this network directly from scratch.
As for G, since the dimension of fake samples here is 3,072, we set $\tt FC1$ and $\tt FC2$ in G to be 2,048 dimensions.

The duration of action instance varies from $<$1s to $>$200s in ActivityNet and thus during evaluation we vary the temporal offset from 10s to 100s. As shown in Table \ref{res_an_table}, \textbf{our approach} significantly outperform the baseline methods again and improves \textbf{TSN w/o ours} which indicates that it also accepts TSN features as input and has the same testing network architecture as \textbf{our approach} but is trained without our proposed methods.

\begin{table}[t]
	\begin{center}
		\caption{AS detection mAP (\%) on ActivityNet when varying the offset threshold}
		\label{res_an_table}
		\begin{tabular}{c|ccc}
			offset threshold (s)     & 10    & 50    & 100   \\ \hline
			Random guess & 0.06  & 0.14  & 0.17  \\
			SceneDetect        & 4.71 &  18.93 & 25.84 \\
			ShotDetect        & 6.10 & 24.35 & 33.76 \\
			TSN w/o ours        & 8.18 & 31.39 & 44.15 \\
			\textbf{Our approach} & \textbf{8.33} & \textbf{33.08} & \textbf{46.97}
		\end{tabular}
	\end{center}
\end{table}

\subsection{Efficiency}
In terms of testing speed, unlike offline detection which evaluates how many frames can be processed per second simultaneously, it is important for ODAS to evaluate the detection delay which is the time duration between the system receives a new video frame and the system outputs the prediction for this frame.
Our model in Fig.~\ref{network} (c) is able to respond within 0.16s on one single Titan X GPU.
Further, our method can maintain similar mAP results even when the striding distance of the input window is increased to 8 frames, thus allowing real-time implementations. 



\section{Conclusion and Future Works}

In this paper, we have proposed a novel Online Detection of Action Start task in a practical setting involving untrimmed, unconstrained videos.
Three training methods have been proposed to specifically improve the capability of ODAS models in detecting action timely and accurately.
Our methods can be applied to any existing video backbone network.
Extensive experiments demonstrate the effectiveness of our approach.
In the future, since the developed methods can address the specific challenges in detecting action starts, it would be interesting to further apply them to help address other online action detection tasks such as per-frame labeling and temporal localization.

\clearpage

\bibliographystyle{splncs03}
\bibliography{egbib}
\end{document}